\documentclass[runningheads]{llncs}

\usepackage{eccv}

\usepackage{eccvabbrv}

\usepackage{graphicx}
\usepackage{booktabs}

\usepackage[accsupp]{axessibility}  

\usepackage{hyperref}

\usepackage{orcidlink}

\usepackage{multirow}
\usepackage{capt-of}
\usepackage{algorithm}
\usepackage{algpseudocode}
\usepackage{wrapfig}

\usepackage[table,dvipsnames]{xcolor}
\usepackage{comment}
\usepackage{capt-of}

\definecolor{green}{rgb}{0, 0.5, 0}
\definecolor{orange}{rgb}{0.8, 0.6, 0.2}
\definecolor{red}{rgb}{1.0, 0.0, 0.0}
\definecolor{teal}{rgb}{0.0, 0.4, 0.4}
\definecolor{purple}{rgb}{0.65,0,0.65}
\definecolor{saffron}{rgb}{0.95,0.75,0.2}
\definecolor{turquoise}{rgb}{0.0,0.5,0.5}
\definecolor{black}{rgb}{0.0, 0.0, 0.0}
\definecolor{gray}{rgb}{0.5, 0.5, 0.5}

\newcommand{\rz}[1]{\textcolor{black}{#1}}

\newcommand{\ourmethod}{PADFormer}

\begin{document}

\title{\ourmethod: Pose-agnostic Anomaly Detection from Sparse View Images} 

\titlerunning{PADFormer}

\author{Ruiqi Wang\inst{1,2}\thanks{Work carried out during internship at Amazon.}\orcidlink{0009-0000-3379-6103} \and
Yiming Qian\inst{1} \and
Fenggen Yu\inst{1}\orcidlink{0000-0003-1591-4668} \and
Yuxuan Lu\inst{3}\orcidlink{0000-0002-8520-0540} \and
Dakuo Wang\inst{3}\orcidlink{0000-0001-9371-9441} \and
Hao Zhang\inst{2}\orcidlink{0000-0003-1991-119X} \and
Jing Huang\inst{1}\orcidlink{0000-0001-8769-9130}
}

\authorrunning{R.Wang et al.}

\institute{Amazon \and
Simon Fraser University \and
Northeastern University
}

\maketitle

\begin{abstract}
Pose-agnostic Anomaly Detection (PAD) remains challenging as anomalies can appear under arbitrary viewpoints, requiring methods to handle significant pose variations. Existing approaches rely on complex 3D reconstruction, which are computationally expensive and require extensive multi-view data. We propose PADFormer, a novel image-space approach that leverages Vision Transformer (ViT) to directly reconstruct anomaly-free versions of query images while preserving pose information. Our key insight is to adapt cross-view masked reconstruction for anomaly detection 
through training exclusively on normal data, combined with dynamic patch selection and spatial alignment mechanisms that enable effective learning from sparse reference views under significant pose variations. During inference, we perform multiple forward passes with different masking patterns to generate an ensemble of anomaly-free reconstructions, ensuring comprehensive coverage of the query image. Anomalies are detected by comparing these reconstructions with the query image. PADFormer achieves state-of-the-art results on the PAD benchmark while maintaining comparable performance on classic few-shot anomaly detection (FSAD) tasks, demonstrating superior efficiency and generalization without requiring 3D reconstruction.
\end{abstract}    
\section{Introduction}
\label{sec:intro}

Unsupervised visual anomaly detection has emerged as a critical task in industrial quality control and manufacturing~\cite{bergmann2019mvtec,bergmann2021mvtec,li2025one,wang2024real, zhu2025real, maack2025pcad}, aiming to identify shape or texture anomalies using only normal samples and representations from pre-trained models~\cite{roth2022towards, defard2021padim}. With the increasing demand for automated inspection of manufactured products and the growing complexity of object shapes and textures, object-centric anomaly detection has attracted significant research interest. However, traditional anomaly detection methods~\cite{defard2021padim, gudovskiy2022cflow, kruse2025multi} operate under a restrictive pose-aligned assumption, where objects are inspected from predefined, fixed viewpoints. This limitation severely constrains their practical applicability, as real-world anomalies may manifest from arbitrary viewing angles and potentially occluded perspectives.

The \emph{pose-agnostic anomaly detection} (PAD) setting~\cite{zhou2023pad} was recently introduced to address this limitation, offering enhanced flexibility by enabling anomaly detection when the query image is captured from an arbitrary, unknown viewpoint. A common practice in PAD is to first reconstruct the target object in 3D from dense anomaly-free input views with ground-truth (GT) poses, then perform cross-comparison between the query image and rendered views from the 3D model after estimating the query pose and aligning it with the 3D reconstruction~\cite{kruse2024splatpose,liu2024splatpose+, yang2025piad,jiang2024igspad}. However, these 3D reconstruction-based methods face critical bottlenecks: they require GT camera poses for hundreds of reference images (often 200+) to build the 3D model, and they incur substantial computational costs due to the 3D reconstruction process. This requirement for dense multi-view data with precise pose annotations poses a significant challenge for real-world deployment, where obtaining comprehensive multi-view coverage with GT poses may be infeasible or prohibitively expensive.
\begin{wrapfigure}{r}{0.45\linewidth}
    \centering
    \vspace{-15pt}
    \includegraphics[width=\linewidth]{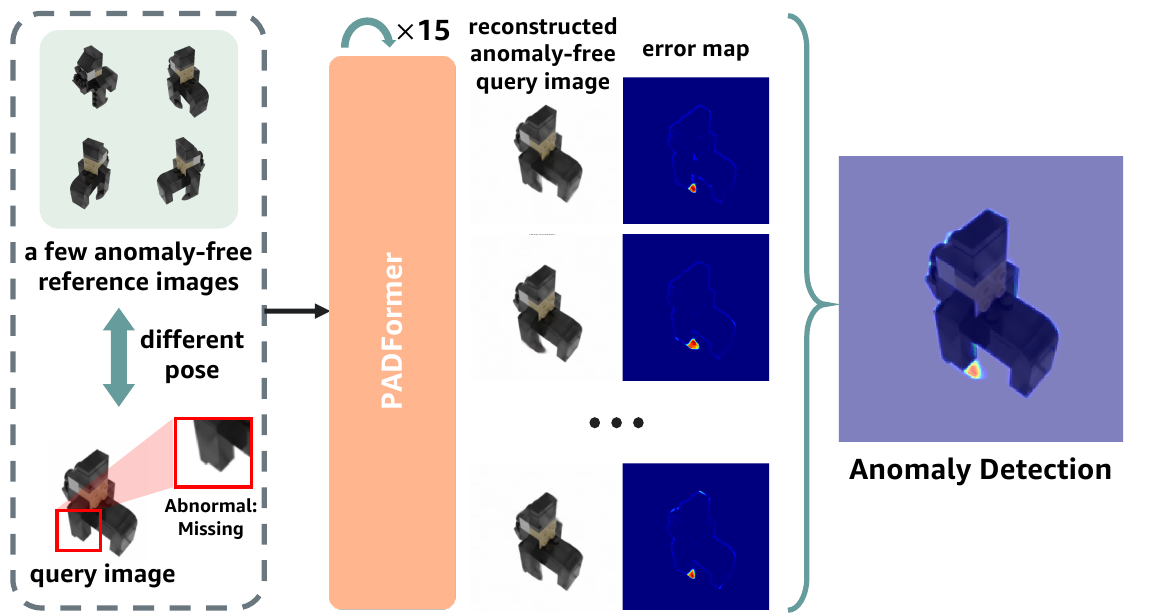}
    \caption{\textbf{Method overview -- } We propose \ourmethod, a novel ViT-based method that directly reconstructs anomaly-free query images from sparse-view references for \emph{pose-agnostic anomaly detection} (PAD). \ourmethod~performs multiple stochastic inference passes to generate diverse reconstructions and error maps, which are averaged to produce final anomaly detection results.}
    \label{fig:teaser}
    \vspace{-15pt}
\end{wrapfigure}
In this paper, we introduce PADFormer, a novel approach that extends pose-agnostic anomaly detection to the \emph{sparse-view} input setting, as illustrated in Fig.~\ref{fig:teaser}. Our key insight is that by leveraging Vision Transformers (ViT)~\cite{dosovitskiy2021image} with a masked reconstruction strategy~\cite{he2022masked}, we can directly reconstruct an \emph{anomaly-free} version of the query \emph{image}, with or without anomalies, while bypassing explicit 3D model acquisition. Anomaly detection and localization can then be accomplished through pixel-wise comparison between the query image and its reconstructed anomaly-free counterpart. Specifically, we apply random masking to the query image, and PADFormer learns to reconstruct the masked patches in an anomaly-free manner by training exclusively on anomaly-free data, naturally preserving the exact pose of the input image. This pipeline significantly reduces the computational complexity associated with dense input view requirements while achieving state-of-the-art performance on PAD benchmarks with sparse views.

To enable effective learning from sparse views, PADFormer incorporates several technical innovations. First, we introduce a dynamic patch selection mechanism with Spatial Transformation Networks (STN)~\cite{jaderberg2015spatial} that efficiently identifies and aligns the most relevant reference patches for each query position, reducing computational overhead while maintaining reconstruction quality. Second, our cross-attention decoder~\cite{vaswani2017attention} leverages learnable anomaly-agnostic tokens that encode both global and local anomaly-free priors, enabling the model to suppress anomalous patterns during reconstruction. Third, we employ a two-stage training strategy to progressively build our model's capability: an initial \emph{within-view} stage learns geometric invariance of object features through self-augmentation, followed by a \emph{cross-view} training that enables generalization across object categories. During inference, we perform multiple reconstruction passes with different masking patterns to ensure comprehensive coverage, computing anomaly scores in CIELAB color space~\cite{sharma2005ciede2000} for robust detection.

Our main contributions are summarized as follows:
\begin{itemize} 
\item We introduce and formalize the sparse-view PAD setting, addressing a critical gap between research assumptions and real-world deployment constraints where dense multi-view data is impractical or unavailable.
\item We propose PADFormer, a novel ViT-based method that directly reconstructs anomaly-free query images in 2D image space, eliminating the need for complex 3D reconstruction while maintaining pose consistency through learned masked reconstruction.
\item Our overall design achieves state-of-the-art performance on PAD benchmarks under sparse view conditions with superior computational efficiency compared to 3D-based methods, while demonstrating strong generalization to classical \emph{few-shot anomaly detection} (FSAD) scenarios. \rz{Note that all existing FSAD benchmarks assume geometric alignment between training and test images, operating under the constraint of similar or identical viewpoints.}
\end{itemize}

\begin{figure}[t]
    \centering
    \includegraphics[width=\linewidth]{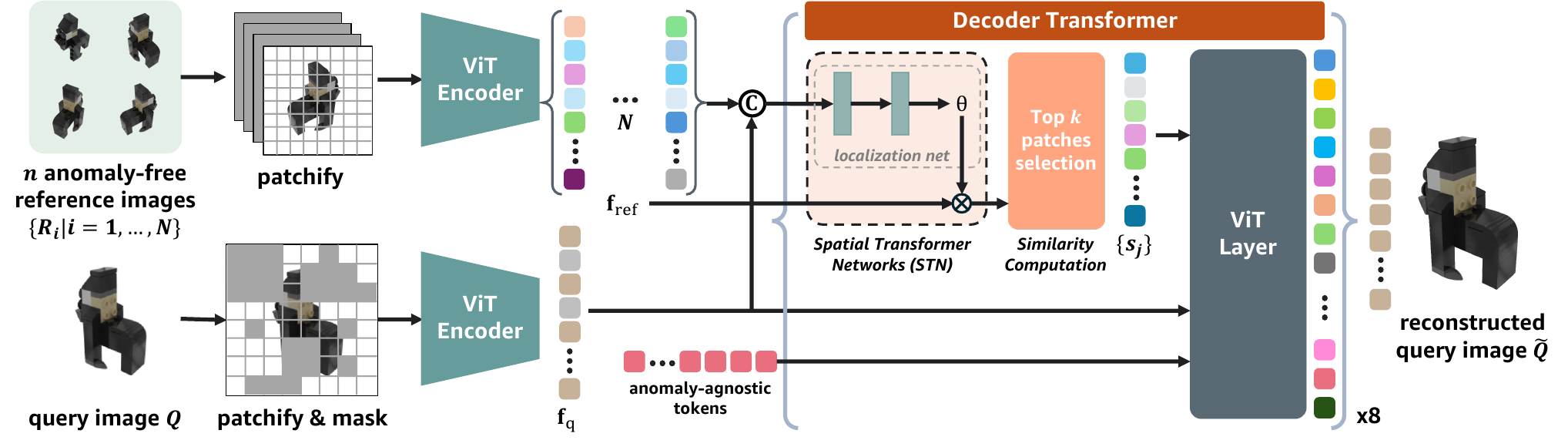}
    \caption{\textbf{\ourmethod~model architecture --} \ourmethod~first patchifies the reference images $\{R_i|i = 1, \ldots, N\}$ into tokens. The query image to be reconstructed $Q$ is randomly masked and also tokenized. Both are processed through two weight-sharing ViT encoders to obtain query feature token $\mathbf{f}_{\text{q}}$ and concatenated reference features token $\mathbf{f}_{\text{ref}}$. A Spatial Transformation Network (STN) takes their concatenation to compute transformation $\boldsymbol{\theta}$, which is applied \textit{only to} $\mathbf{f}_{\text{ref}}$ via bilinear sampling for alignment, producing $\tilde{\mathbf{f}}_{\text{ref}}$. A similarity computation module (SIM) then selects top-$k$ most similar aligned reference patches $\{\mathbf{s}_j\}$ for each query position. In the decoder, query features token $\mathbf{f}_{\text{q}}$ attend to anomaly-agnostic tokens and selected reference patches $\{\mathbf{s}_j\}$ through cross-attention across 8 ViT layers. A linear projection maps the output to pixel values, which are unpatchified to produce the reconstructed query image $\tilde{Q}$.}
    \label{fig:method}
    \vspace{-1em}
\end{figure}
\section{Related Works}
\label{sec:rw}

\subsection{Anomaly Detection}
\paragraph{Reference-free Image Anomaly Detection.}
Classical reference-free image anomaly detection framework trains a customized model for each class to detect anomalies without requiring normal reference images during testing. Self-supervised learning methods~\cite{golan2018deep, sohn2021learning, reiss2021panda} learn normal patterns through pretext tasks, while hybrid approaches~\cite{roth2022towards, defard2021padim} integrate feature matching with distribution modeling. Transformer-based methods~\cite{luo2025inp, you2022unified, jin2025swin, yang2024unsupervised, nafez2025patchguard} leverage self-attention mechanisms to capture long-range dependencies for anomaly localization. However, these methods assume fixed object poses with similar viewpoints. In contrast, our work addresses pose-agnostic anomaly detection where objects exhibit significant pose variations.

\paragraph{Pose-agnostic Anomaly Detection (PAD).} 
Current approaches to PAD predominantly rely on explicit 3D geometric reconstruction pipelines. Zhou et al.~\cite{zhou2023pad} introduced this problem alongside the Multi-pose Anomaly Detection (MAD) dataset, which contains 20 object categories with over 200 views each. Their OmniposeAD method leverages Neural Radiance Fields (NeRF) to encode anomaly-free objects from diverse viewpoints, performs coarse-to-fine pose estimation on query images, and detects anomalies by comparing reconstructed normal references with queries. To improve reconstruction efficiency and quality,  SplatPose~\cite{kruse2024splatpose} and SplatPose++~\cite{liu2024splatpose+} adopt 3D Gaussian Splatting for faster training and inference. More recently, PIAD~\cite{yang2025piad} extends the framework to jointly handle pose and illumination variations. Despite their effectiveness, these 3D reconstruction-based approaches require extensive multi-view training data with pose annotations, suffer from reconstruction artifacts, and need test-time pose optimization for each query. In contrast, our method enables efficient PAD by directly reconstructing anomaly-free query images from few reference views, eliminating 3D reconstruction and test-time optimization.

\paragraph{Few-shot Anomaly Detection (FSAD).} 
Current FSAD methods can be categorized into three groups: (1) Memory-Bank and Embedding Methods, such as PatchCore~\cite{roth2022towards} and PaDiM~\cite{defard2021padim}, which leverage coreset subsampling to define representative patch-level features from few normal samples; (2) Vision-Language Models (VLMs), which exploit pretrained models through prompt learning~\cite{li2024promptad, jeong2023winclip, gu2024anomalygpt, ma2025aa, zhou2023anomalyclip, cao2024adaclip, qu2024vcp, xu2025towards, sadikaj2025multiads, luo2025exploring} or in-context residual learning~\cite{zhu2024toward} for robust cross-domain generalization; and (3) Reconstruction and Registration Methods~\cite{fang2023fastrecon, huang2022registration, huang2024few,beizaee2025correcting}, which employ 2D spatial transformers or diffusion models to align query features with support set features. UniVAD~\cite{gu2025univad} represents the current state-of-the-art training-free FSAD method,  leveraging contextual component clustering and multi-level matching to achieve superior cross-domain generalization. 
While these methods excel at data efficiency, they fundamentally lack the capability to handle unconstrained 3D pose variations. Their 2D alignment mechanisms cannot synthesize the novel appearances required by arbitrary viewpoint changes, limiting their applicability to real-world scenarios where test-time viewpoints differ from reference views.

Our method bridges PAD and FSAD by addressing their complementary limitations: we achieve pose-agnostic anomaly detection without requiring extensive multi-view data, pose annotations, or 3D reconstruction pipelines. By learning to reconstruct anomaly-free query images directly from few-shot references, we simultaneously achieve data efficiency and pose robustness.

\subsection{Masked Image Modeling}
Masked Image Modeling (MIM)\cite{devlin2019bert, he2022masked} has emerged as a powerful self-supervised learning paradigm that trains models to predict masked image regions from visible context. Pioneering works include BEiT\cite{bao2021beit}, which reconstructs discrete visual tokens, and MAE~\cite{he2022masked}, which introduced a high masking ratio (75\%) with an asymmetric encoder-decoder design. Subsequent methods~\cite{xie2022simmim, zhouimage, gupta2023siamese} advanced the field through simplified architectures, self-distillation, and siamese designs. MAEDAY~\cite{schwartz2024maeday} pioneered MIM for anomaly detection, leveraging the insight that anomalous regions are harder to reconstruct. Recent works~\cite{irshad2024nerf, rajasegaran2025gaussian} have explored MIM for 3D-aware tasks. However, MAEDAY and similar reconstruction-based methods~\cite{zavrtanik2021draem, zavrtanik2021riad} focus on viewpoint-aligned images and cannot handle unconstrained pose variations in PAD scenarios.

Extending MIM to multi-view scenarios, CroCo~\cite{weinzaepfel2022croco} introduces cross-view completion where a partially masked input image is reconstructed using both its visible content and a second unmasked reference image from a different viewpoint. Unlike single-view MIM, cross-view completion resolves reconstruction ambiguity by reasoning about scene geometry across views. However, CroCo trains on arbitrary scenes for general 3D vision tasks. Our work differs by adapting cross-view MIM for anomaly detection through training exclusively on anomaly-free data. This enables synthesizing anomaly-free versions of query images under novel viewpoints, where anomalies are replaced with plausible normal content rather than faithfully reconstructed.

\section{Method}
\label{sec:method}
We begin by formulating the problem statement in Sec.~\ref{sec:ps}, followed by a detailed description of our transformer-based model architecture in Sec.~\ref{sec:model}. Finally, we elaborate on the training and inference procedures in Sec.~\ref{sec:training}.

\subsection{Problem Statement}
\label{sec:ps}

Given a test query image $Q_{\text{test}}$ (normal or abnormal) and a set of $N$ anomaly-free reference images $\{R_i\}_{i=1}^{N}$ from object $o_j$, the goal is to detect and localize anomalies in $Q_{\text{test}}$. Both $Q_{\text{test}}$ and $\{R_i\}_{i=1}^{N}$ are captured under unknown and different camera poses. In addition, the reference set $\{R_i\}_{i=1}^{N}$ is small and sparse, providing limited viewpoint coverage of the object $o_j$.

Previous Few-Shot Anomaly Detection (FSAD) methods assume the query and reference images share aligned viewpoints, limiting their ability to detect defects from unseen angles. Conversely, early \emph{pose-agnostic anomaly detection} (PAD) approaches require dense reference sets and per-object training, which reduces their practical applicability. This work addresses both limitations by tackling anomaly detection in a pose-agnostic manner with only a sparse set of reference images, eliminating the need for viewpoint alignment or extensive training data.

\subsection{Model Architecture}
\label{sec:model}
As shown in Fig.~\ref{fig:method}, \ourmethod~uses an end-to-end transformer model to learn the anomaly-free query image reconstruction. In this subsection, we first describe the key components of our \ourmethod, then provide details of the training and inference for localization of the anomalies.
\paragraph{Encoder.}
\ourmethod~starts by tokenizing the input images. Given a query image $Q$ and $N$ reference images $\{R_i|i=1,\ldots,N\}$, we first patchify them into non-overlapping patches of size $p \times p$ and linearly project each patch to obtain tokens of dimension $d$. Fixed sinusoidal positional embeddings are added to all tokens. The query and reference images are then processed through the same Vision Transformer (ViT) encoder with shared weights. The key difference lies in the masking strategy: similar to Siamese MAE~\cite{gupta2023siamese}, we adopt an asymmetric masking approach where we randomly mask a ratio (40\%) of patches in the query image, keeping only the visible patches as input to the encoder, while all reference images are encoded without any masking. The encoder consists of $L$ standard transformer blocks with multi-head self-attention and feed-forward layers. After encoding, we obtain the masked query features $\mathbf{f}_q \in \mathbb{R}^{M_v \times d}$ where $M_v$ is the number of visible patches, and complete reference features $\{\mathbf{f}_{r_i} \in \mathbb{R}^{M \times d} | i=1,...,N\}$ where $M$ is the total number of patches per image. These encoded features serve as the foundation for subsequent patch-level matching and reconstruction. This asymmetric design forces the model to aggregate information from the intact reference images to reconstruct the masked query regions, allowing it to learn anomaly-free reconstruction patterns.

\paragraph{Dynamic Patch Selection.}
To enable efficient and accurate cross-attention between query and reference images, we employ a dynamic patch selection mechanism that identifies the most relevant reference patches for each query position. This approach serves two purposes: (1) improving reconstruction quality by focusing on the most informative reference patches, and (2) reducing computational cost by limiting the number of patches used in cross-attention. Let $\mathbf{f}_{r_i} \in \mathbb{R}^{M \times d}$ denote the encoded features of the $i$-th reference image, where $M$ is the number of patches per image. Before measuring patch similarity, we align each reference feature grid to the query feature grid with a Spatial Transformation Network (STN)~\cite{jaderberg2015spatial}. The STN is used only to transform reference features; the query features remain fixed and define the target coordinate frame.

The STN operates on patch-level features through the following pipeline. Given visible query features $\mathbf{f}_\text{q} \in \mathbb{R}^{M_v \times d}$ after masking, we first fill masked positions with learnable embeddings and restore the original patch order, producing a complete query grid $\hat{\mathbf{f}}_\text{q} \in \mathbb{R}^{M \times d}$. For each reference image $R_i$, $\hat{\mathbf{f}}_\text{q}$ and $\mathbf{f}_{r_i}$ are reshaped from token sequences into 2D feature maps of shape $[B,d,h,w]$, where $h \times w=M$. Their channel-wise concatenation is passed to a lightweight localization network with two convolutional layers (kernel sizes $7 \times 7$ and $5 \times 5$, with 32 and 10 output channels), each followed by $2 \times 2$ max pooling, and fully connected layers that predict a 6-dimensional affine transformation matrix $\boldsymbol{\theta}_i$. This affine transform is applied only to the reference feature map: we generate a sampling grid from $\boldsymbol{\theta}_i$ and use bilinear interpolation in feature space to obtain the aligned reference features $\tilde{\mathbf{f}}_{r_i}$. After applying this process to all $N$ references, the aligned features are concatenated into $\tilde{\mathbf{f}}_\text{ref}=\{\tilde{\mathbf{f}}_{r_i}\}_{i=1}^{N} \in \mathbb{R}^{(N \cdot M) \times d}$. This patch-level geometric alignment reduces pose-induced mismatch before nearest-patch retrieval.

We then measure the relevance between query and aligned reference patches using a hybrid similarity score that combines feature and spatial cues. The feature similarity between query patch $j$ and aligned reference patch $l$ is computed via normalized cosine similarity: 
\begin{equation}
    S_{\text{feat}}^{jl} = \langle \frac{\hat{\mathbf{f}}_\text{q}^j}{\|\hat{\mathbf{f}}_\text{q}^j\|}, \frac{\tilde{\mathbf{f}}_\text{ref}^l}{\|\tilde{\mathbf{f}}_\text{ref}^l\|} \rangle
\end{equation}
When the query and reference images have matching resolutions, we additionally incorporate spatial proximity: 
\begin{equation}
    S_{\text{spatial}}^{jl} = \exp\left(-\frac{\|\mathbf{p}_\text{q}^j - \mathbf{p}_\text{ref}^l\|_2}{\sigma}\right)
\end{equation}
where $\mathbf{p}_\text{q}^j$ and $\mathbf{p}_\text{ref}^l$ are the 2D grid positions of patches in their original spatial arrangements, and $\sigma=2.0$ controls spatial locality. The final similarity is:
\begin{equation}
    S^{jl} = (1-w) S_{\text{feat}}^{jl} + w S_{\text{spatial}}^{jl}
\end{equation}
where $w=0.3$ balances feature matching and spatial locality. Based on these scores, we select the top-$k$ most similar aligned reference patches for each query position, denoted as $\mathbf{s}_j \in \mathbb{R}^{k \times d}$ (with $k=10$ by default). Thus, each query token attends only to a compact, pose-aligned set of reference candidates rather than all $N \cdot M$ reference patches.

\paragraph{Cross-Attention Decoder.}
The decoder reconstructs the masked query patches through 8 ViT layers, where each layer combines query context, learned anomaly-free priors, and retrieved reference evidence. The encoded query features $\mathbf{f}_\text{q}$ and aligned reference features $\tilde{\mathbf{f}}_\text{ref}$ are first projected to the decoder dimension $d'$ via separate linear layers. The masked query positions are filled with learnable mask embeddings, and the full query sequence is reordered to the original patch positions using the unmasking indices from the encoder. Decoder positional embeddings are then added to all query tokens.

To provide reconstruction priors that do not depend on a specific anomaly type, we introduce learnable anomaly-agnostic tokens as a trainable memory bank. This memory contains (1) 32 global tokens $\mathbf{T}_\text{global} \in \mathbb{R}^{32 \times d'}$ shared across all positions to capture common normal appearance patterns, and (2) position-indexed local tokens $\mathbf{T}_\text{local} \in \mathbb{R}^{M \times 1 \times d'}$, where each patch position has one associated token for location-aware normal priors. We flatten the local tokens into $\mathbf{T}_\text{local}^\text{flat} \in \mathbb{R}^{M \times d'}$ and expose all local tokens to every query position during cross-attention. This all-to-all access is important under large pose changes: the most useful normal prior for a query patch may come from a different nominal grid location, while the positional embeddings still preserve each token's location identity.

Each decoder layer proceeds in two steps. First, self-attention among the query tokens propagates visible-query context and updates the masked-token representations. Second, for each query position $j$, the decoder performs cross-attention over a position-specific key/value set formed by concatenating three sources: the global anomaly-agnostic tokens $\mathbf{T}_\text{global}$, the flattened local tokens $\mathbf{T}_\text{local}^\text{flat}$, and the selected aligned reference patches $\mathbf{s}_j$. In this interaction, the query token decides how much to use global normal priors, location-aware priors, and concrete visual evidence from the reference images. After 8 layers and final layer normalization, a linear projection maps the decoded features to pixel values, which are then unpatchified to produce the reconstructed query image $\tilde{Q}$.

\subsection{Training \& Inference}
\begin{wrapfigure}{r}{0.4\linewidth}
    \centering
    \vspace{-20pt}
    \includegraphics[width=\linewidth]{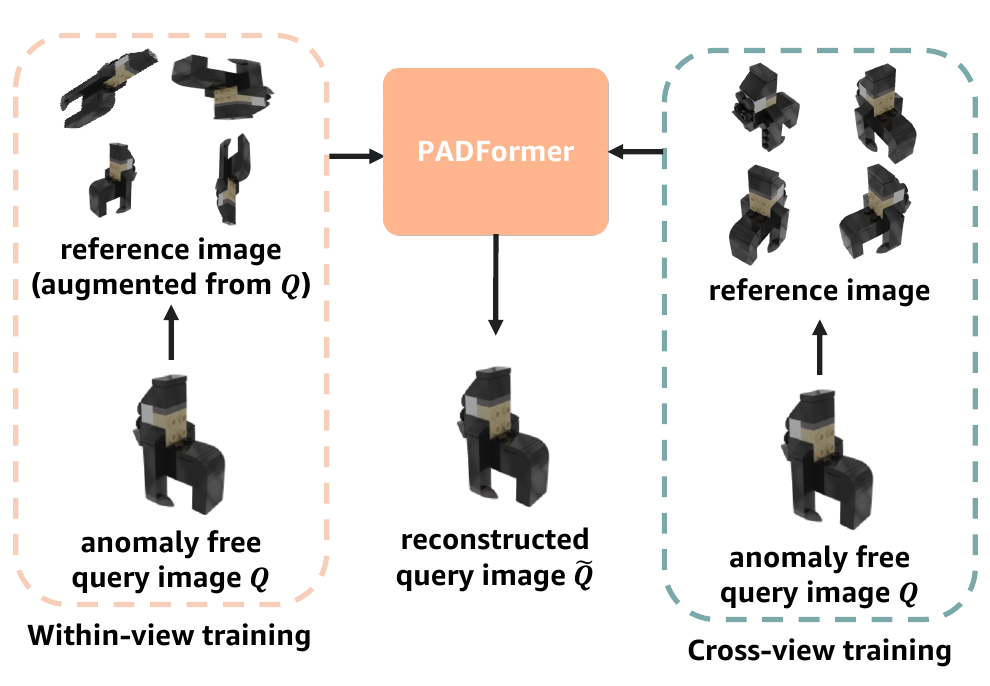}
    \vspace{-2em}
    \caption{\textbf{Within-view pre-training.} An augmented version of the query image is used as reference.}
    \label{fig:withintrain}
    \vspace{-25pt}
\end{wrapfigure}
\label{sec:training}
\paragraph{Training.}
\ourmethod~aims to learn to reconstruct anomaly-free query images based on few-shot reference images. We train our model in two stages to facilitate this objective. During training, both reference images $\{R_i|i=1, \dots, N\}$ and query image $Q$ are anomaly-free.

\begin{itemize}
    \item \textbf{Within-view training:} For each query image $Q$, we generate reference images by applying random rotation and deformation transformations to the query image $Q$ itself. The model is then trained to reconstruct the original query image $Q$ from these self-augmented references, as shown in Fig.~\ref{fig:withintrain}. This strategy enables the model to learn geometric invariance and robust feature representations while focusing on intrinsic object properties, laying a solid foundation for cross-object generalization in the subsequent stage.
\end{itemize}

\begin{itemize}
    \item \textbf{Cross-view training: } In this stage, we train a unified \ourmethod~model across all object categories $O$ in the dataset simultaneously. Unlike the within-view stage where references are derived from the query itself, here the model learns to reconstruct query images using reference images from the same object category but different views. This cross-view training enables the model to generalize beyond instance-specific features and learn category-level representations that are robust to intra-class variations.
\end{itemize}

For both training stages, following prior works~\cite{hong2023lrm, jin2024lvsm}, we train \ourmethod~with MSE and perceptual loss~\cite{johnson2016perceptual}:
\begin{equation}
    L = \text{MSE}(Q, \tilde{Q}) + \lambda \cdot \text{Perceptual}(Q, \tilde{Q})
\end{equation}
where $\tilde{Q}$ denotes the reconstructed query image and $\lambda$ is the weight for balancing the perceptual loss.

\paragraph{Inference.}
During inference, given a test query image $Q_\text{test}$, we randomly sample $N=4$ anomaly-free reference images from the training set. We perform $I=15$ reconstruction passes with different random masking patterns. With mask ratio $r=0.4$, the probability that any patch is masked at least once is $1-(1-r)^{I} \approx 99.95\%$, providing near-complete spatial coverage. For each reconstruction pass $i$, we first upsample the reconstructed image $\hat{Q}_i$ to match the original resolution for sharper anomaly localization. We then compute pixel-wise reconstruction errors using perceptual color differences in CIELAB space~\cite{sharma2005ciede2000}. Specifically, we apply Gaussian filtering ($\sigma=1.4$, kernel size 7) to each LAB channel of the query image to reduce high-frequency noise, then compute squared differences with the unfiltered reconstruction:

\begin{equation}
E_i(x,y) = \sum_{c \in \{L,a,b\}} \left(\mathcal{G}_\sigma\left(c(Q_\text{test})\right) - c(\hat{Q}_i)\right)^2
\end{equation}
where $\mathcal{G}_\sigma$ denotes Gaussian filtering and $c$ indexes the LAB channels. The $I$ error maps are averaged: $\bar{E}(x,y) = \frac{1}{I}\sum_{i=1}^{I}E_i(x,y)$. To distinguish genuine anomalies from consistent reconstruction artifacts, we compute the variance $\text{Var}(x,y) = \frac{1}{I}\sum_{i=1}^{I}(E_i - \bar{E})^2$ across reconstructions and retain only high-variance regions via a consistency mask $\mathcal{M}(x,y) = \mathbb{1}[\text{Var}(x,y) > P_{50}(\text{Var})]$, where $P_{50}$ is the 50th percentile. This variance-based filtering also provides a natural defense against false positives arising from rare-but-normal patterns such as batch variations or illumination drift. Since such patterns are part of the normal data distribution, the model reconstructs them consistently across different masking passes, resulting in low variance. They are therefore suppressed by the consistency mask, unlike genuine anomalies which the model fails to reconstruct consistently. The final anomaly score map is:
\begin{equation}
\mathcal{E}(x,y) = \bar{E}(x,y) \cdot \mathcal{M}(x,y)
\end{equation}
The image-level anomaly score is $S = \max(\mathcal{E})$, providing a single scalar score for classification.
\section{Experiments}
\label{sec:exp}
In this section, we describe our experimental setup and datasets (Sec.~\ref{sec:data}), introduce our model training details (Sec. ~\ref{sec:training-details}), report evaluation results (Sec.~\ref{sec:eval}) and perform an ablation study (Sec.~\ref{sec:ablation}).
\begin{table}[t]
    \caption{\textbf{Anomaly detection; PAD task --} Quantitative comparison of image-level AUROC, pixel-level AUROC, and AUPRO on MAD-SIM and PIAD datasets. Best results are in \textbf{bold} and second-best are \underline{underlined}.}
    \label{tab:mad-quan}
    \centering
    \resizebox{\linewidth}{!}{
    \scriptsize
    \begin{tabular}{llcccccccc}
    \toprule
     \multirow{2}{*}{\textbf{Setup}} & \multirow{2}{*}{\textbf{Method}} & \multicolumn{3}{c}{\textbf{MAD-SIM}} & & \multicolumn{3}{c}{\textbf{PIAD (synt+real)}}\\
     \cmidrule{3-5}\cmidrule{7-9}
     & & \multicolumn{1}{c}{image-AUROC} & \multicolumn{1}{c}{pixel-AUROC} & \multicolumn{1}{c}{AUPRO} & & \multicolumn{1}{c}{image-AUROC} & \multicolumn{1}{c}{pixel-AUROC} & \multicolumn{1}{c}{AUPRO}\\
     \midrule
     \multirow{6}{*}{2-shot}& OmniAD~\cite{zhou2023pad} & 48.2 & 80.1 & 68.2 && 47.1 & 77.5 & 66.7 \\
     & SplatPose~\cite{kruse2024splatpose}& 48.3 & 81.4 & 69.1 && 48.1 & 77.2 & 66.9\\
     & PIAD~\cite{yang2025piad} & 49.0 & \underline{82.2} & \underline{70.4} & & 48.1 & \underline{78.5} & \underline{69.2}\\
     & PromptAD~\cite{li2024promptad} & 56.4 & 78.9 & 67.3 && 53.2 & 70.8 & 60.2\\
     & UniVAD~\cite{gu2025univad} & \underline{57.9} & 80.4 & 68.8 && \underline{54.3} & 71.9 & 60.7\\
     \rowcolor{blue!10!white} & \textbf{PADFormer} & \textbf{78.8} & \textbf{89.2} & \textbf{76.5} & & \textbf{75.6} & \textbf{86.3} & \textbf{74.1}\\
     \midrule
     \multirow{6}{*}{4-shot}& OmniAD~\cite{zhou2023pad} & 50.6 & 84.6 & 72.9 & & 47.5 & 78.5 & 68.4\\
     & SplatPose~\cite{kruse2024splatpose} & 52.2 & \underline{86.7} & \underline{74.6} & & 48.8 & 78.9 & 68.8\\
     & PIAD~\cite{yang2025piad} & 52.4 & 81.1 & 68.4 & & 54.7 & \underline{84.4} & \underline{72.3}\\
     & PromptAD~\cite{li2024promptad} & \underline{60.3} & 78.9 & 67.5 & & 58.8 & 77.5 & 66.9\\
     & UniVAD~\cite{gu2025univad} & 58.4 & 80.0 & 67.9 & & \underline{59.2} & 79.4 & 69.8\\
     \rowcolor{blue!10!white} &\textbf{PADFormer} & \textbf{81.3} & \textbf{92.9} & \textbf{86.7} && \textbf{80.7} & \textbf{91.2} & \textbf{79.4}\\
     \midrule
     \multirow{4}{*}{10-shot}& OmniAD~\cite{zhou2023pad} & 51.0 & 86.8 & 74.9 & & 64.2 & 83.5 & 71.1\\
     & SplatPose~\cite{kruse2024splatpose} & 56.4 & 90.2 & 83.8 && 67.4 & 87.3 & 76.5\\
     & PIAD~\cite{yang2025piad} & \underline{58.2} & \underline{91.8} & \underline{85.5} && \underline{78.3} & \underline{90.5} & \underline{78.6}\\
     \rowcolor{blue!10!white}& \textbf{PADFormer}& \textbf{85.6} & \textbf{94.4} & \textbf{88.1} & & \textbf{84.3} & \textbf{93.4} & \textbf{82.5} \\ 
     \midrule
     \multirow{4}{*}{All-shot}& OmniAD~\cite{zhou2023pad} & 91.9 & 98.4 & 92.3 & & 79.1 & 95.9 & 88.3\\
     & SplatPose~\cite{kruse2024splatpose} & 94.9 & \underline{99.0} & \underline{94.1} & & 82.6 & 97.7 & 93.1\\
     & PIAD~\cite{yang2025piad} & \underline{97.4} & \textbf{99.5} & \textbf{95.4} & & \underline{94.7} & \textbf{99.0} & \textbf{95.2}\\
     \rowcolor{blue!10!white}& \textbf{PADFormer} & \textbf{97.8} & 98.9 & 93.9 & & \textbf{95.4} & \underline{98.3} & \underline{93.7}\\
     \bottomrule
    \end{tabular}
    }
\end{table}
\subsection{Datasets}
\label{sec:data}
\paragraph{PAD Dataset.} We use the MAD-SIM\footnotemark{}\footnotetext{We did not use MAD-Real since OmniAD~\cite{zhou2023pad} strongly recommends using MAD-Sim exclusively to explore the PAD problem in their GitHub repository.} dataset~\cite{zhou2023pad} and PIAD dataset~\cite{yang2025piad} to train \ourmethod~separately. For each anomaly-free image in the training set, we randomly select $N \in \{2, 4, 10\}$ other anomaly-free images from the same training set to serve as reference images. During testing, we evaluate on the test sets of both datasets. For each test image, we randomly select $N \in \{2, 4, 10, \text{all}\}$ anomaly-free images from the training set as references.
\paragraph{FSAD Datasets.} We use the MVTec-AD dataset~\cite{bergmann2019mvtec} and ViSA dataset~\cite{zou2022spot} to train \ourmethod~separately. For each anomaly-free image in the training set, we randomly select $N \in \{1, 2, 4\}$ other anomaly-free images from the same training set to serve as reference images. During testing, we evaluate on the test sets of both datasets. For each test image, we randomly select $N \in \{1, 2, 4\}$ anomaly-free images from the training set as references. We follow the standard evaluation protocol established by prior FSAD works~\cite{roth2022towards,jeong2023winclip,gu2025univad}, which report results exclusively on $\{1, 2, 4\}$-shot settings, as these low-shot scenarios are the primary focus of few-shot anomaly detection research.
\subsection{Training Details}
\label{sec:training-details}
We train separate models for each dataset, with all categories trained together in a unified model per dataset. We use a pretrained MAE~\cite{he2022masked} encoder as our backbone. All input images are resized to 224$\times$224. The training process consists of two stages: we first apply within-view training on the model for 50 epochs, followed by a second stage of cross-view training for 250 epochs. For the within-view stage, we apply random rotations and elastic deformations to generate reference images from each query. We employ a cosine learning rate schedule with a peak learning rate of 4e-4 and a warmup period of 2500 steps. All experiments are conducted on 4 NVIDIA A10 GPUs. We train separate models for each value of $N \in \{2, 4, 10, \text{all}\}$ for PAD datasets and $N \in \{1, 2, 4\}$ for FSAD datasets to evaluate the impact of the number of reference images on performance.

\subsection{Comparison to Baselines}
\label{sec:eval}
\begin{figure*}[t]
    \centering
    \includegraphics[width=\linewidth]{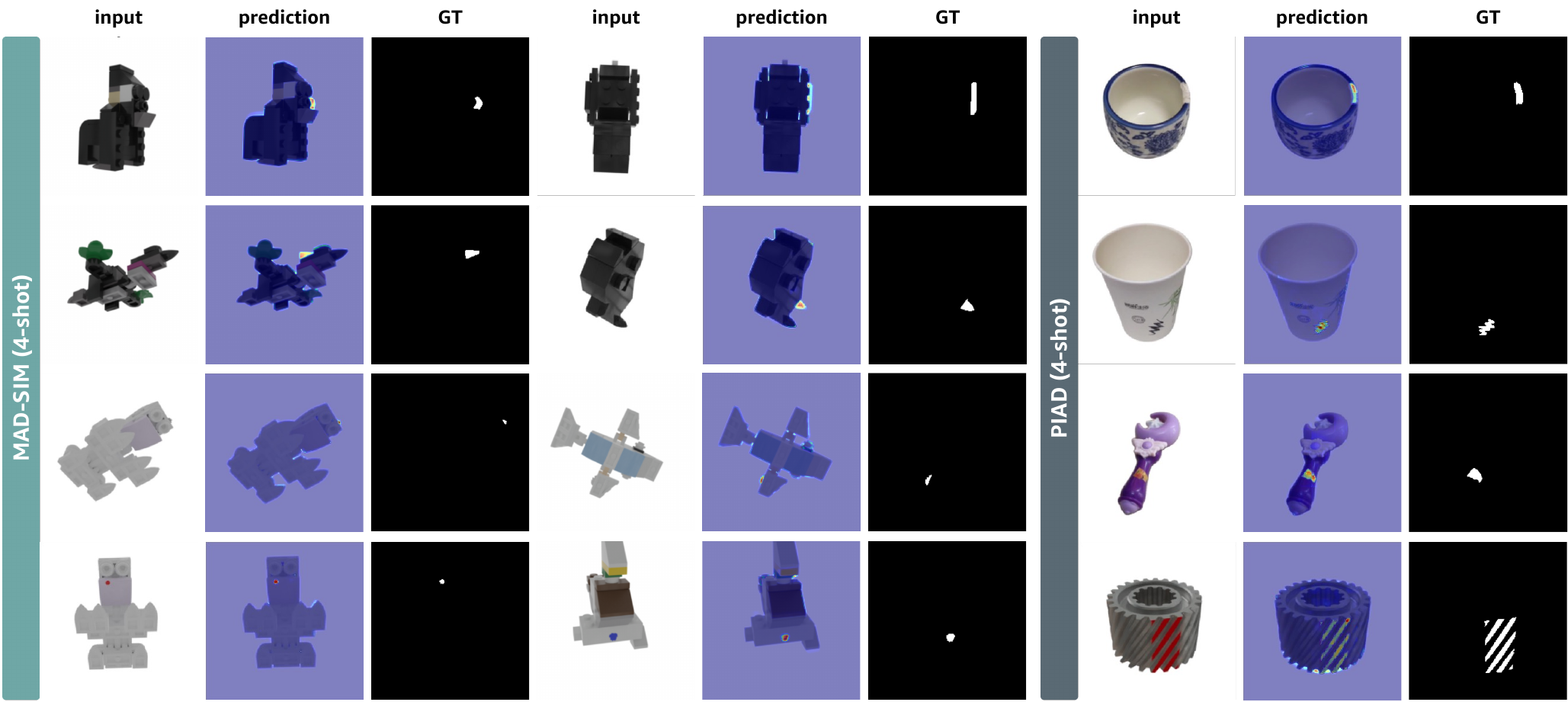}
    \caption{\textbf{Qualitative results on PAD benchmarks --} On the left we visualize the results on MAD-SIM, with missing parts (row 1, 2), burrs (row 3) and stains (row 4). On the right we visualize results on PIAD. See supplementary materials for more qualitative comparisons.}
    \vspace{-1em}
    \label{fig:qual-pad}
\end{figure*}

Following previous works~\cite{bergmann2019mvtec, zhou2023pad}, we evaluate the performance of \ourmethod~using two standard metrics for anomaly detection: Area Under the Receiver Operating Characteristic curve (AUROC) at image-level and pixel-level. Pixel-AUROC measures the model's ability to localize anomalous regions at the pixel level, while image-AUROC evaluates the accuracy of binary classification for determining whether an entire image contains an anomaly. Both metrics range from 0 to 100, with higher values indicating better performance.
\begin{table*}[t]
    \caption{\textbf{Anomaly detection; FSAD task -- } Quantitative comparison of image and pixel-level AUROC on MVTec-AD and VisA dataset. }
    \label{tab:fsad-quan}
    \centering
    \resizebox{\linewidth}{!}{
    \small
    \begin{tabular}{llccccc}
    \toprule
     \multirow{2}{*}{\textbf{Setup}} & \multirow{2}{*}{\textbf{Method}} & \multicolumn{2}{c}{\textbf{MVTecAD}} & & \multicolumn{2}{c}{\textbf{VisA}}\\
     \cmidrule{3-4}\cmidrule{6-7}
     & & \multicolumn{1}{c}{image-level (AUROC)} & \multicolumn{1}{c}{pixel-level (AUROC)} & & \multicolumn{1}{c}{image-level (AUROC)} & \multicolumn{1}{c}{pixel-level (AUROC)}\\
     \midrule
     \multirow{5}{*}{1-shot}& PatchCore~\cite{roth2022towards} & 63.7 & 83.9 & & 58.9 & 76.7\\
     & WinCLIP~\cite{jeong2023winclip} & \underline{92.8} & 92.4 & & 83.1 & 94.6\\
     & PromptAD~\cite{li2024promptad} & 86.3 & 91.8 & & 80.8 & \underline{96.3} \\
     & UniVAD~\cite{gu2025univad} & \textbf{94.8} & \textbf{95.7} & & \textbf{87.1} & \textbf{97.2} \\
     \rowcolor{blue!10!white}& \textbf{PADFormer} & \underline{92.8} & \underline{92.0} & & \underline{83.6} & 95.0\\
     \midrule
     \multirow{5}{*}{2-shot}& PatchCore~\cite{roth2022towards}& 72.4 & 89.6 & & 60.2 & 82.4\\
     & WinCLIP~\cite{jeong2023winclip} & \underline{92.7} & \underline{92.4} & & 83.7 & 95.1\\
     & PromptAD~\cite{li2024promptad} & 89.2 & 92.2 & & 84.3 & \underline{96.9} \\
     & UniVAD~\cite{gu2025univad}& \textbf{96.4} & \textbf{96.0} && \textbf{89.3} & \textbf{97.5}\\
     \rowcolor{blue!10!white}& \textbf{PADFormer} & \underline{92.8} & 92.0 & &\underline{84.6} & 95.2\\
     \midrule
     \multirow{5}{*}{4-shot}& PatchCore~\cite{roth2022towards} & 74.9 & 92.6 & & 62.6 & 85.4\\
     & WinCLIP~\cite{jeong2023winclip} & 94.0 & 92.9 & & 84.1 & 95.2\\
     & PromptAD~\cite{li2024promptad} & 90.6 & 92.4 & & 85.7 & \underline{97.2} \\
     & UniVAD~\cite{gu2025univad} & \textbf{98.6} & \textbf{96.1} && \textbf{91.7} & \textbf{97.8}\\
     \rowcolor{blue!10!white}& \textbf{PADFormer} & \underline{94.7} & \underline{93.2} & &\underline{86.6} & 95.4\\
     \bottomrule
    \end{tabular}
    }
\end{table*}
\paragraph{PAD Results.}
We compare \ourmethod~with OmniAD~\cite{zhou2023pad}, SplatPose~\cite{kruse2024splatpose}, and PIAD~\cite{yang2025piad} on PAD task, as shown in Tab.~\ref{tab:mad-quan}. Our method demonstrates significant advantages in few-shot setting, outperforming the second-best method by over 20\% on image-level AUROC and 8\% on pixel-level AUROC in the 2-shot setup. Similar trends are observed in 4-shot and 10-shot setup, validating the effectiveness of \ourmethod~in PAD task with sparse reference images. In the all-shot setting, \ourmethod~maintains the best image-level AUROC on both benchmarks, while achieving competitive pixel-level results. This marginal pixel-level performance gap is attributed to inherent reconstruction errors. Notably, all compared methods perform per-object training in this setting, wheras \ourmethod~is trained jointly on all object categories, demonstrating its generalization ability.  Beyond detection performance, \ourmethod~also offers superior computational efficiency for inference, as detailed in the supplementary materials. Qualitative results on both benchmarks under 4-shot setup are shown in Fig.~\ref{fig:qual-pad}. \ourmethod~successfully detects both fine-grained defects (e.g., stains and burrs) and relative large defects (e.g., missing parts) on MAD-SIM and PIAD, demonstrating robust performance across diverse viewpoints and anomaly types.

\begin{figure}[t]
    \centering
    \includegraphics[width=\linewidth]{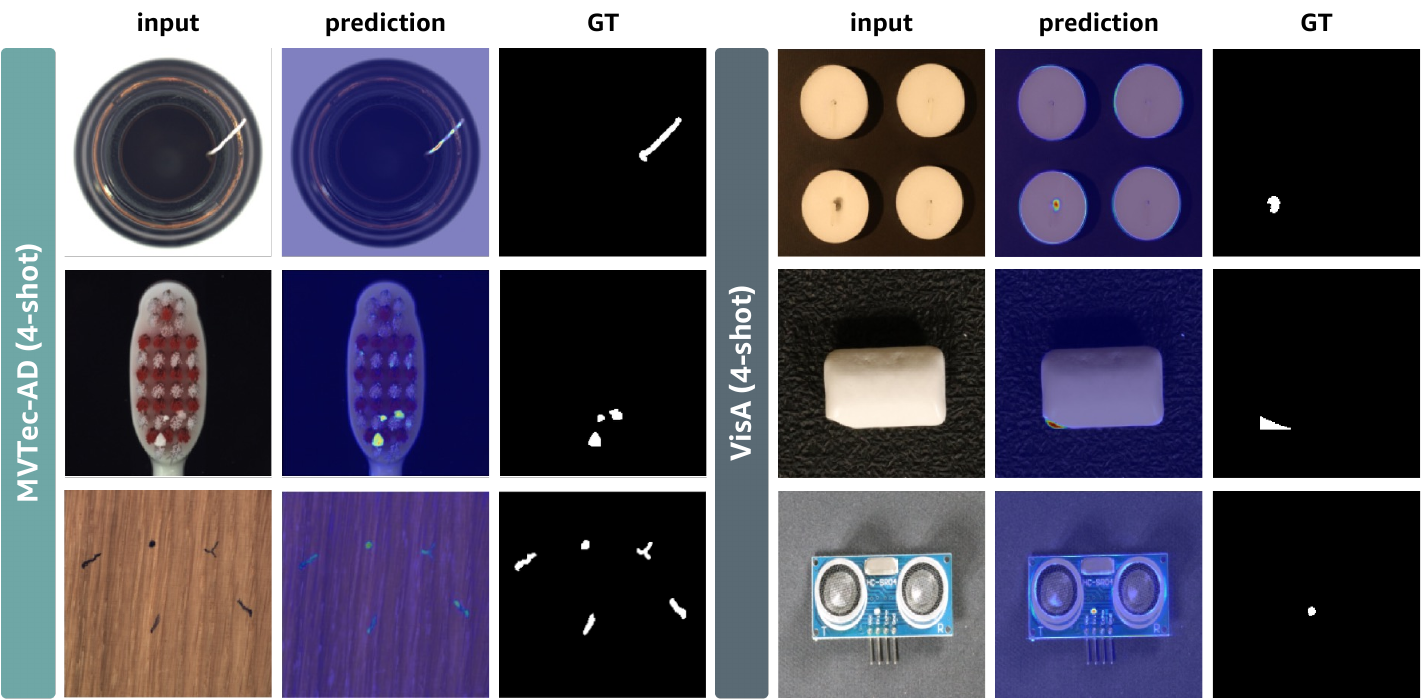}
    \caption{\textbf{Qualitative results on FSAD benchmarks:} MVTec-AD (L) and VisA (R).}
    \label{fig:qual-fsad}
    \vspace{-1em}
\end{figure}

\paragraph{FSAD Results.}
We further compare \ourmethod~with PatchCore\cite{roth2022towards}, WinCLIP~\cite{jeong2023winclip}, PromptAD~\cite{li2024promptad} and UniVAD~\cite{gu2025univad}. As shown in Tab.~\ref{tab:fsad-quan}, PADFormer achieves competitive performance on the general FSAD task. Importantly, \ourmethod~is designed for PAD, where query images capture arbitrary viewpoints of the object, while general FSAD assumes geometric alignment between query and reference views, which is a fundamentally easier setting that stronger FSAD baselines such as UniVAD are specifically optimized for, leveraging large foundational models and alignment-specific architectures. Despite this task mismatch, our method demonstrates strong generalization without relying on such large foundational models, making it more lightweight and self-contained. Crucially, no existing FSAD method achieves comparable performance on PAD benchmarks, confirming that PADFormer addresses a distinctly harder problem. Fig.~\ref{fig:qual-fsad} shows qualitative results on MVTec-AD~\cite{bergmann2019mvtec} and VisA~\cite{zou2022spot} under 4-shot setup. \ourmethod~effectively handles challenging cases including defects in densely crowded regions (e.g., toothbrush bristles) and subtle anomalies such as tiny stains on PCBs.


\begin{table}[t]
    \begin{minipage}{.499\linewidth}
      \centering
      \caption{\textbf{Ablation -- } on model architecture, including Spatial Transformation Network (STN), dynamic patch selection (DPS) and anomaly-agnostic token (AAT).}
      \resizebox{0.9\columnwidth}{!}{
        \begin{tabular}{cccccc}
        \toprule
        \multicolumn{3}{c}{\textbf{Modules in PADFormer}} & &\multicolumn{2}{c}{\textbf{MAD (4-shot)}}\\
        \cmidrule{1-3}\cmidrule{5-6}
        STN & DPS & AAT & & image-AUROC & pixel-AUROC \\
        \midrule
        - & - & - & & 76.8 & 80.2 \\
        - & \checkmark & - & & 78.2 & 85.4\\
        \checkmark & \checkmark & - & & 79.1 & 89.2 \\
        \checkmark & \checkmark & \checkmark & & \textbf{81.3} & \textbf{92.9}\\
        \bottomrule
        \end{tabular}
        }
        \label{tab:ablation}
    \end{minipage}%
    \hspace{0.02\textwidth}
    \begin{minipage}{.499\linewidth}
      \centering
        \caption{\textbf{Ablation -- } on anomaly detection error metric.}
        \resizebox{0.9\columnwidth}{!}{
        \begin{tabular}{lccc}
        \toprule
        \multirow{2}{*}{\textbf{Error Metric}} &\multicolumn{2}{c}{\textbf{MAD (4-shot)}} & \multirow{2}{*}{\textbf{Time (s)} $\downarrow$}\\
        \cmidrule{2-3}
         & image-AUROC & pixel-AUROC \\
         \midrule
         RGB & 78.8 & 86.4 & \textbf{0.6}\\
         LAB & \underline{81.3} & \textbf{92.9} & \underline{0.8}\\
         ViT feature & \textbf{83.5} & \underline{91.3} & 2.1 \\
        \bottomrule
        \end{tabular}
        \label{tab:ablation-inference}
        }
    \end{minipage} 
\end{table}
\subsection{Ablation Studies}
\label{sec:ablation}
\paragraph{Model Architecture.}
As shown in Tab.~\ref{tab:ablation}, we evaluate the effectiveness of our model designs. The baseline without any proposed modules uses a standard cross-attention transformer decoder, which is structurally analogous to CroCo~\cite{weinzaepfel2022croco} adapted for anomaly detection through training on anomaly-free data only. This baseline achieves 76.8\% and 80.2\% on image-level and pixel-level AUROC, respectively. Adding the Dynamic Patch Selection (DPS) module for top-$k$ reference patches selection in cross-attention improves performance to 78.2\% and 85.4\%, demonstrating that selecting the most relevant patches is more effective than attending to all reference patches. Incorporating the Spatial Transformation Network (STN) to spatially align reference features before similarity computation further boosts the metrics to 79.1\% and 89.2\%, as it enables more accurate patch matching across different views. Finally, adding the Anomaly-agnostic Tokens (AAT)—32 global and $M$ position-indexed local learnable tokens—achieves the best performance of 81.3\% and 92.9\%. Each component contributes meaningfully, with the complete PADFormer achieving 4.5\% and 12.7\% improvements over the CroCo-style baseline.

\paragraph{Error Evaluation Method.}
In Tab.~\ref{tab:ablation-inference}, we ablate the anomaly scoring metric across three representation spaces: RGB space, LAB space and feature space extracted from intermediate transformer layers. While RGB space offers the fastest computation, it achieves significantly lower performance in both image-AUROC and pixel-AUROC due to its lack of perceptual uniformity. Feature space attains the highest image-level performance, but at approximately 3 times higher computational cost. LAB space achieves the best pixel-level localization while maintaining near-optimal efficiency, only marginally slower than RGB computation. Our LAB-based metric applies Gaussian filtering to the query image channels before computing differences, which reduces sensitivity to high-frequency noise and edge artifacts. The variance-based consistency filtering (retaining top 50\% variance regions) contributes an additional 2.4\% improvement in image-AUROC by suppressing systematic reconstruction errors that appear consistently across all masking patterns.

\section{Conclusion}
We present PADFormer, a novel Vision Transformer-based approach for \emph{pose-agnostic anomaly detection} (PAD) under sparse-view settings. Our method addresses a critical limitation in existing approaches: the ability to perform few-shot PAD without requiring pose information from either reference or query images, while maintaining low computational cost. By directly reconstructing anomaly-free query images in 2D space through learned masked reconstruction, PADFormer eliminates the computationally expensive 3D reconstruction pipeline employed by prior methods, yet achieves superior pose consistency. Our method achieves superior performance compared to existing PAD approaches across few-shot settings, demonstrating exceptional sample efficiency in handling arbitrary viewpoints. PADFormer exhibits strong robustness and generalization across diverse object categories, while competitive performance on general FSAD benchmarks, where geometric alignment is assumed, further validates the versatility of our learned representations.

Despite these strengths, PADFormer has limitations. The fixed patch size may suboptimally handle large anomalies, and multiple inference passes (15 runs) are needed for comprehensive coverage. Future work could explore adaptive patch mechanisms and intelligent masking strategies to improve efficiency, as well as lightweight pose estimation to enhance performance under extreme viewpoint variations.


\section*{Acknowledgements}
We sincerely thank all reviewers and Area Chair for carefully reading our paper, supplementary materials, and rebuttal. We appreciate their insightful suggestions and encouraging comments. We also thank Xi Liu and Weiwei Sun from Amazon for helpful discussions and valuable comments; Zhengqing Wang from Simon Fraser University for sharing helpful knowledge and experience with feed-forward transformers; Mingrui Zhao from Simon Fraser University for proofreading. 

%
%
\bibliographystyle{splncs04}
\bibliography{main}

\clearpage
\setcounter{page}{1}
\setcounter{section}{0}
\renewcommand{\thesection}{\Alph{section}}
{\centering
\Large
\vspace{0.5em}\textbf{Supplementary Material} \\}
\vspace{2em}
The supplementary document provides (1) details of dynamic patch selection and cross-attention decoder in \ourmethod{} in Sec.~\ref{sec:supp-details}; (2) additional ablation studies in Sec.~\ref{sec:supp-ablation}; (3) additional quantitative results in Sec.~\ref{sec:supp-quan}; and (3) additional qualitative results in Sec.~\ref{sec:supp-qual}. 

\section{Implementation details of \ourmethod}
\label{sec:supp-details}
In this section, we provide implementation details of dynamic patch selection and cross attention deployed in \ourmethod. A pseudocode is presented in Alg.~\ref{supp:alg}.

\paragraph{Dynamic Patch Selection.}
The Spatial Transformation Network (STN)~\cite{jaderberg2015spatial} employed for reference alignment consists of a localization network with a specific two-stage convolutional architecture. The first convolutional layer uses a $7\times7$ kernel with 32 output channels, followed by $2\times2$ max pooling with stride 2 and ReLU activation. The second convolutional layer applies a $5\times5$ kernel with 10 output channels, again followed by $2\times2$ max pooling and ReLU activation. The output feature map is then flattened and processed through two fully-connected layers: the first maps to a hidden dimension of 32 with ReLU activation, and the second outputs 6 parameters representing the $2 \times 3$ affine transformation matrix. To ensure training stability, the final fully-connected layer is initialized with zero weights and biases set to $[1, 0, 0, 0, 1, 0]$, corresponding to an identity transformation. This initialization prevents the network from applying arbitrary spatial distortions at the beginning of training.

The hybrid similarity computation adaptively switches between two modes depending on image resolution compatibility. When query and reference images have matching dimensions (i.e., $M_q = M_{\text{ref}}$), both feature and spatial similarities are computed. The 2D grid positions $\mathbf{p}_q^j$ and $\mathbf{p}_{\text{ref}}^l$ are obtained by converting linear patch indices into 2D coordinates: for a patch at index $i$, its position is computed as $(\lfloor i / \sqrt{M} \rfloor, i \bmod \sqrt{M})$. When resolutions differ, the spatial similarity term is omitted and only feature-based cosine similarity is used for patch selection. The top-$k$ selection is performed independently for each query position using PyTorch's \texttt{torch.topk} operation, which efficiently identifies the $k$ patches with highest similarity scores. The selected patch indices are then used with \texttt{torch.gather} to extract the corresponding features from the aligned reference set, producing position-specific reference patches $\mathbf{s}_j \in \mathbb{R}^{k \times d'}$ for each query position $j$.

\paragraph{Decoder with cross attention.}
The decoder begins by projecting both query and reference features from encoder dimension $d=768$ (for base model) or $d=1024$ (for large model) to decoder dimension $d'=512$ using separate linear projection layers. Reference features from all $N$ reference images are concatenated along the sequence dimension after removing their CLS tokens, resulting in a combined reference set of size $(N \cdot M) \times d'$. For masked query positions, learnable mask token embeddings are inserted and the sequence is reordered according to the restoration indices produced by the encoder's random masking operation. Fixed 2D sinusoidal positional embeddings are then added to all decoder tokens.

\begin{algorithm}[t]
\caption{Cross-Attention Decoder with Dynamic Reference Patches Selection}
\label{supp:alg}
\label{alg:decoder}
\begin{algorithmic}[1]
\Require Encoded query $\mathbf{f}_q \in \mathbb{R}^{(L+1) \times d}$, reference features $\{\mathbf{f}_{\text{ref}}^{(n)}\}_{n=1}^N$, restoration indices $\mathbf{ids}_{\text{restore}}$
\Ensure Reconstructed patches $\hat{\mathbf{x}} \in \mathbb{R}^{M \times (p^2 \cdot 3)}$

\State \textbf{// Initialization}
\State $\mathbf{h}^{(0)} \leftarrow \text{Linear}_{d \rightarrow d'}(\mathbf{f}_q)$
\State $\mathbf{r} \leftarrow \text{Concat}(\{\text{Linear}_{d \rightarrow d'}(\mathbf{f}_{\text{ref}}^{(n)}[1:M])\}_{n=1}^N)$ 
\State $\mathbf{h}^{(0)}[1:M] \leftarrow \text{Unmask}(\mathbf{h}^{(0)}, \mathbf{m}, \mathbf{ids}_{\text{restore}})$
\State $\mathbf{h}^{(0)} \leftarrow \mathbf{h}^{(0)} + \mathbf{PE}_{\text{dec}}$

\For{$i = 1$ to $8$}
    \State $\mathbf{h}^{(i)} \leftarrow \text{LayerNorm}(\mathbf{h}^{(i-1)} + \text{SelfAttn}(\mathbf{h}^{(i-1)}))$
    
    \State \textbf{// Dynamic patch selection}
    \State $\mathbf{q} \leftarrow \mathbf{h}^{(i)}[1:M]$
    \State $\tilde{\mathbf{r}} \leftarrow \text{STN}(\mathbf{q}, \mathbf{r})$
    
    \For{$j = 1$ to $M$}
        \State $\mathbf{S}_{\text{feat}}^{j} \leftarrow \text{CosineSim}(\mathbf{q}^j, \tilde{\mathbf{r}})$
        \If{$M_q = M_{\text{ref}}$}
            \State $\mathbf{S}_{\text{spatial}}^{j} \leftarrow \exp(-\|\mathbf{p}_q^j - \mathbf{p}_{\text{ref}}\|_2 / \sigma)$
            \State $\mathbf{S}^{j} \leftarrow (1-w) \mathbf{S}_{\text{feat}}^{j} + w \mathbf{S}_{\text{spatial}}^{j}$
        \Else
            \State $\mathbf{S}^{j} \leftarrow \mathbf{S}_{\text{feat}}^{j}$
        \EndIf
        \State $\mathbf{s}_j \leftarrow \tilde{\mathbf{r}}[\text{TopK}(\mathbf{S}^{j}, k)]$
    \EndFor
    
    \State \textbf{// Construct keys and values}
    \State $K = V \leftarrow \text{Concat}([\mathbf{g}, \{\mathbf{l}_j\}_{j=1}^M, \{\mathbf{s}_j\}_{j=1}^M])$
    
    \State \textbf{// Cross-attention}
    \State $\mathbf{h}^{(i)} \leftarrow \text{LayerNorm}(\mathbf{h}^{(i)} + \text{CrossAttn}(\mathbf{h}^{(i)}, K, V))$
\EndFor

\State $\hat{\mathbf{x}} \leftarrow \text{Linear}_{d' \rightarrow p^2 \cdot 3}(\text{LayerNorm}(\mathbf{h}^{(8)})[1:M])$

\noindent\Return $\hat{\mathbf{x}}$
\end{algorithmic}
\end{algorithm}

The decoder begins by projecting both query and reference features from encoder dimension $d=768$ (for base model) or $d=1024$ (for large model) to decoder dimension $d'=512$ using separate linear projection layers. Reference features from all $N$ reference images are concatenated along the sequence dimension after removing their CLS tokens, resulting in a combined reference set of size $(N \cdot M) \times d'$. For masked query positions, learnable mask token embeddings are inserted and the sequence is reordered according to the restoration indices produced by the encoder's random masking operation. Fixed 2D sinusoidal positional embeddings are then added to all decoder tokens.

At each decoder layer, the cross-attention mechanism constructs keys and values by concatenating three components in sequence: (1) the 32 global tokens, (2) all $M$ local tokens (one per position) flattened to dimension $M \times d'$, and (3) the selected reference patches flattened to dimension $(M \cdot k) \times d'$. This results in key and value matrices of total size $(32 + M + M \cdot k) \times d'$. With default settings ($M=196$ for $224\times224$ images with $16\times16$ patches, $k=10$), each decoder token attends over $32 + 196 + 1960 = 2188$ total tokens. However, due to the position-specific selection, each query position $j$ effectively focuses on only $32 + 1 + 10 = 43$ relevant tokens: the global priors, its corresponding local prior, and its 10 selected reference patches. The cross-attention uses 16 attention heads with dimension $d'/16 = 32$ per head, following the standard multi-head attention mechanism with \texttt{batch\_first=True} configuration. After cross-attention, a residual connection and layer normalization are applied before proceeding to the next layer.

The dynamic patch selection mechanism is re-executed at each decoder layer using the updated decoder states as queries. This allows the model to progressively refine which reference patches are most relevant as the reconstruction evolves through the decoder depth. Early layers may focus on coarse structural alignment, while later layers can select patches for fine-grained texture details. After all 8 decoder layers, the final layer normalization is applied, the CLS token is removed, and a linear projection layer maps each of the $M$ patch representations from dimension $d'$ to dimension $p^2 \cdot 3 = 768$ (for patch size $p=16$), corresponding to the RGB values of all pixels within each patch. These projected features are then unpatchified by reshaping and rearranging to form the final reconstructed image.

\section{Additional ablation studies}
\label{sec:supp-ablation}
\begin{table}[t]
    \centering
    \caption{\textbf{Efficiency comparison -- }inference time (sec.). (*) Per-object training required methods only compute the pose alignment and anomaly detection time. }
    \begin{tabular}{lccc}
    \toprule
    \textbf{Method} & \textbf{4-shot} & \textbf{10-shot} & \textbf{all-shot}\\
    \midrule
    OmniAD~\cite{zhou2023pad} & 39$^*$ & 38$^*$ & 33$^*$\\
    SplatPose~\cite{kruse2024splatpose} & 2.96$^*$& 2.87$^*$ &2.53$^*$ \\
    PIAD~\cite{yang2025piad} & 3.18$^*$ & 3.14$^*$ & 2.84$^*$\\
    \rowcolor{blue!10!white}\textbf{\ourmethod} & \textbf{0.74} & \textbf{0.83} & \textbf{1.75} \\
    \bottomrule
    \end{tabular}
    \label{tab:supp-time}
\end{table}
We first report the time efficiency of \ourmethod{} and other methods for PAD task in Tab.~\ref{tab:supp-time}. PADFormer demonstrates substantial efficiency advantages over existing methods across all few-shot settings. While competing methods require per-object training (4.5 hours per object for OmniAD and at least 5 minutes per object for 3DGS-based methods) and only report pose alignment and anomaly detection time, they still incur significantly higher computational costs.  Notably, these methods exhibit reduced inference time as the number of reference images increases, which reflects easier pose alignment when the reconstructed NeRF or 3D Gaussian Splatting models are of higher quality. However, this improvement comes at the cost of substantially increased training time with additional references. In contrast, PADFormer achieves inference times of only 0.74s, 0.83s, and 1.75s for 4-shot, 10-shot, and all-shot settings respectively, representing speedups of 4-53 times over the fastest baseline. Importantly, PADFormer's scalability is enabled by its top-$k$ selection mechanism in the decoder, which maintains computational efficiency regardless of reference set size by attending only to the most relevant patches during inference, eliminating the need for costly per-object training while achieving superior speed.

We further provide additional ablation studies on training strategy, masking ratio and number of inference runs separately. 

Tab.~\ref{tab:supp-training} demonstrates the effectiveness of our two-stage training strategy. Within-view training alone yields poor results because it only exposes the model to self-augmented references from the query image, limiting it to learning instance-specific geometric transformations without capturing category-level appearance variations. Cross-view training alone achieves near-optimal performance by learning from diverse object instances, but lacks the geometric robustness provided by within-view augmentation. Our two-stage approach achieves the best results by combining their complementary strengths: within-view training first establishes geometric invariance and robust feature representations, while cross-view training subsequently builds category-level understanding that generalizes across instances. This sequential learning proves more effective than either stage alone, validating that geometric invariance serves as a crucial foundation for learning generalizable anomaly-free representations. The performance gap between cross-view alone and two-stage training, though smaller, confirms that the within-view stage provides non-trivial benefits by conditioning the model to handle geometric variations before tackling cross-instance generalization.
\begin{table}[t]
    \centering
    \caption{\textbf{Ablation studies -- }two-stage training conducted on 4-shot MAD-SIM.}
    \begin{tabular}{lc}
    \toprule
    \textbf{Setting} & \textbf{image/pixel-level AUROC} \\
    \midrule
    Within-view training only & 71.8 / 82.4 \\
    Cross-view training only & 79.2 / 90.1\\
    \rowcolor{blue!10!white}\textbf{Two stage training} & \textbf{81.3 / 92.9} \\
    \bottomrule
    \end{tabular}

    \label{tab:supp-training}
\end{table}
\begin{table}[ht]
    \centering
    \caption{\textbf{Ablation studies -- }masking ratio and inference runs conducted on 4-shot MAD-SIM}
    \begin{tabular}{lc||lc}
    \toprule
    \textbf{Masking Ratio} & \textbf{AUROC (i./p.)} & \textbf{Inference Runs} & \textbf{AUROC (i./p.)}\\
    \midrule
    0.3 & 80.1 / 89.4 & 5 & 72.4 / 81.2\\
    0.4 & \textbf{81.3 / 92.9} & 10 & 78.2 / 87.5\\
    0.6 & 80.7 / 88.6 & 15 & \textbf{81.3 / 92.9} \\
    0.75 & 75.2 / 85.3 & 20 & 81.3 / 92.8 \\ 
    \bottomrule
    \end{tabular}

    \label{tab:supp-params}
\end{table}
\begin{figure*}[t]
    \centering
    \includegraphics[width=\linewidth]{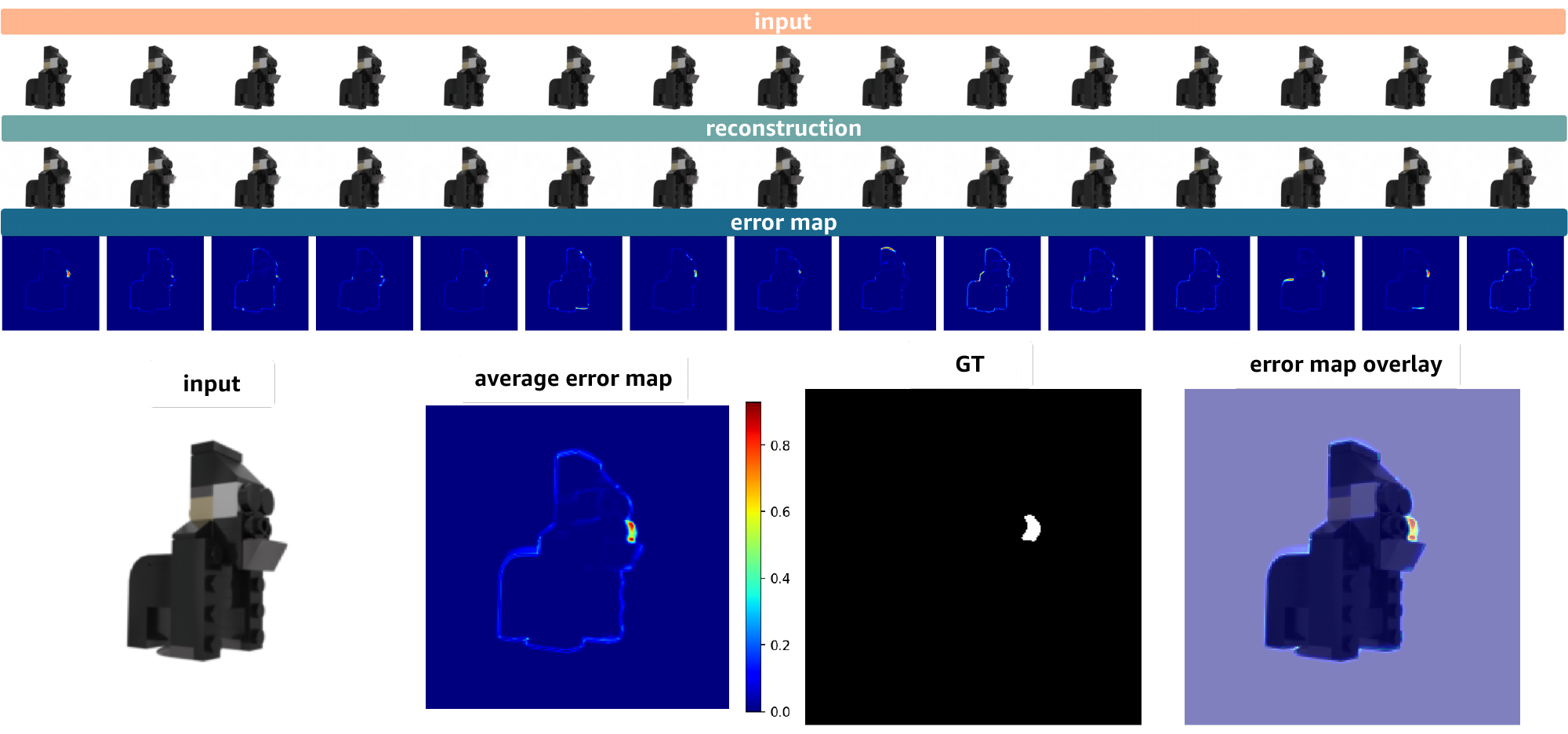}
    \caption{\textbf{Qualitative results - }15 inference runs and their average error map on MAD-Sim with 4-shot setup.}
    \label{fig:supp-qual-1}
\end{figure*}

Tab.~\ref{tab:supp-params} analyzes two critical hyperparameters: masking ratio and number of inference runs. The masking ratio of 0.4 achieves optimal performance, balancing reconstruction difficulty with sufficient context. Lower ratios (0.3) provide inadequate coverage, while higher ratios (0.6/0.75) remove too much context, both degrading performance. 

The number of inference runs shows monotonic improvement from 5 to 20 runs, as more passes ensure comprehensive spatial coverage with different masking patterns. However, performance plateaus beyond 15 runs (81.3\%/92.9\%), with 20 runs providing only marginal gains. We select 15 runs as our default setting, achieving near-optimal performance while maintaining practical efficiency. This multi-run strategy is essential for PADFormer's success: it guarantees that each image region is reconstructed multiple times under different contexts, enabling robust anomaly detection through ensemble averaging and variance-based filtering of reconstruction artifacts.

\section{Additional quantitative results}
\label{sec:supp-quan}
We include experiments required during the review progress and additional results presented in the rebuttal to this section. 

\paragraph{Dynamic patch selection details.}
We analyze $k$ sensitivity under 4-shot MAD-SIM. Smaller $k$ risks missing the correct matching patch under large pose variations when STN alignment is imperfect, hurting pixel-AUROC. Larger $k$ introduces more low-similarity candidates into cross-attention, slightly risking reconstruction toward non-normal content, while increasing cross-attention FLOPs by $\sim$1.9$\times$. $k$=10 achieves the optimal accuracy-efficiency trade-off:
\begin{table}[h]
\vspace{-1em}
\centering
\begin{tabular}{lcccc}
\toprule
$k$ & image-AUROC $\uparrow$ & pixel-AUROC $\uparrow$ & Inference time (s) $\downarrow$ \\
\midrule
5   & 80.6 & 92.3 & 1.70 \\
10  & 81.3 & 92.9 & 1.75 \\
20  & 81.5 & 92.8 & 2.1 \\
\bottomrule
\end{tabular}
\vspace{-1em}
\end{table}

\paragraph{Reference selection analysis.}
We evaluate three reference strategies under 4-shot MAD-SIM. Best-case (4 neighboring views) improves performance as small pose gaps make STN alignment well-posed. Worst-case (1 neighboring + 3 opposite views) degrades performance since 2D affine is limited in representing large out-of-plane rotations. Nevertheless, degradation remains graceful, demonstrating PADFormer's robustness under poor reference coverage. This also provides additional evidence for STN's effectiveness: well-posed alignment directly translates to better performance.
\begin{table}[h]
\vspace{-1em}
\centering
\begin{tabular}{lccc}
\toprule
Metric & Best (4 neighbors) & Random (current) & Worst (1 neighbor) \\
\midrule
image-AUROC & 84.2 & 81.3 & 76.2 \\
pixel-AUROC & 93.7 & 92.9 & 89.1 \\
\bottomrule
\end{tabular}
\vspace{-1em}
\end{table}

\paragraph{MAD-Real evaluation.}
We evaluate PADFormer on MAD-Real under 4-shot setting. However, MAD-Real places objects on a highly reflective white marble background with complex patterns and lighting, making anomaly-free image reconstruction inherently harder and thus degrading detection quality. After segmenting the object from the background as preprocessing, results improve notably. Remarkably, PADFormer with only 4 shots achieves competitive results against OmniAD with 50 shots; see table below.

\begin{table}[h]
\vspace{-1em}
\centering
\scriptsize
\resizebox{\columnwidth}{!}{
\begin{tabular}{lcccc}
\toprule
Metric & PADFormer (4-shot) & PADFormer (4-shot, seg.) & OmniAD (50-shot)\\
\midrule
image-AUROC & 76.4 & 80.4 & 81.0 \\
pixel-AUROC & 84.2 & 91.8 & 95.5 \\
\bottomrule
\end{tabular}
}
\end{table}

\section{Additional qualitative results}
\label{sec:supp-qual}
\begin{figure*}[t]
    \centering
    \includegraphics[width=\linewidth]{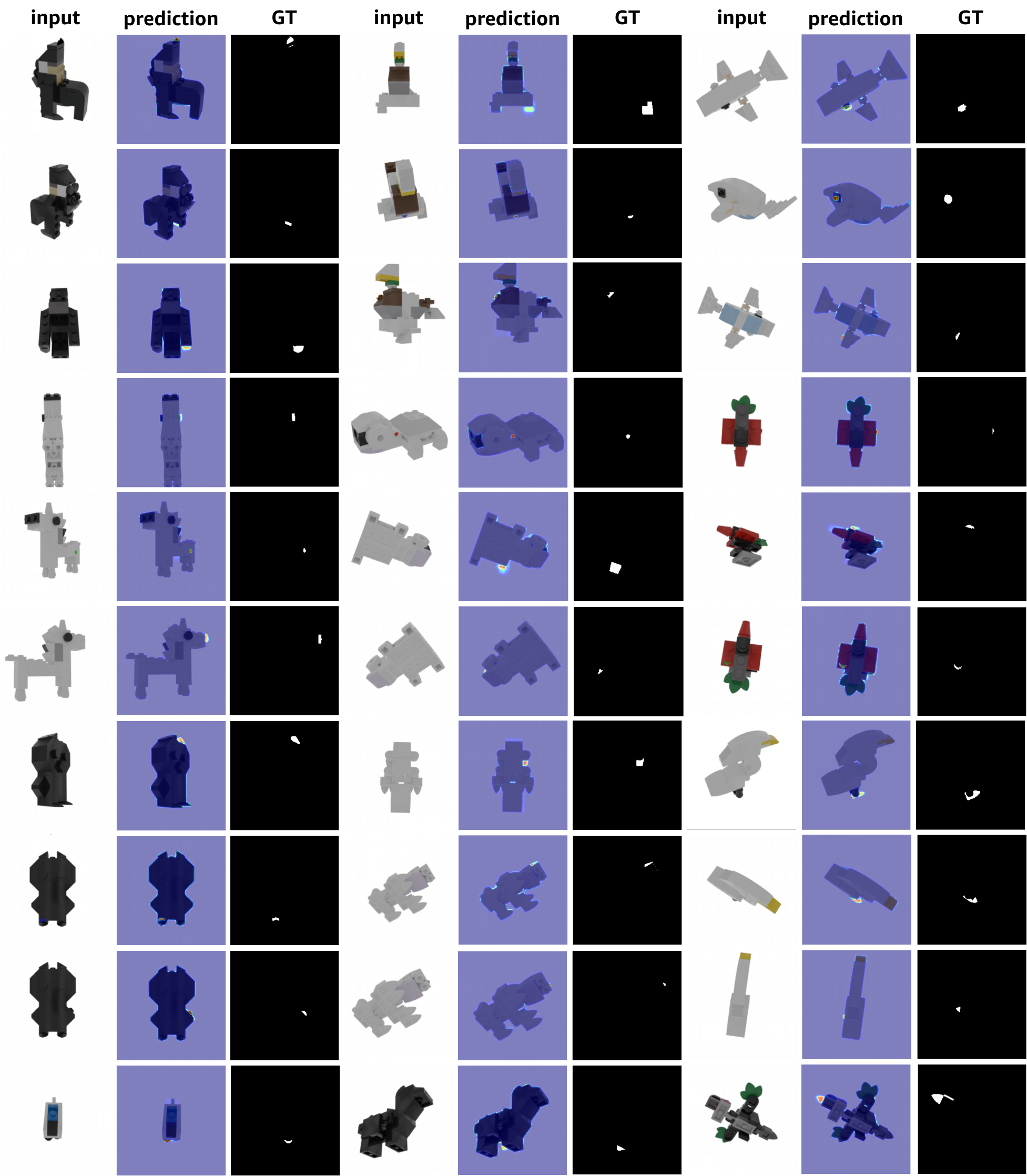}
    \caption{\textbf{Qualitative results - }MAD-Sim 4-shot setup. Our method can detect anomaly including missing, stains, and burrs effectively. }
    \label{fig:supp-qual-mad}
\end{figure*}

\begin{figure*}[t]
    \centering
    \includegraphics[width=\linewidth]{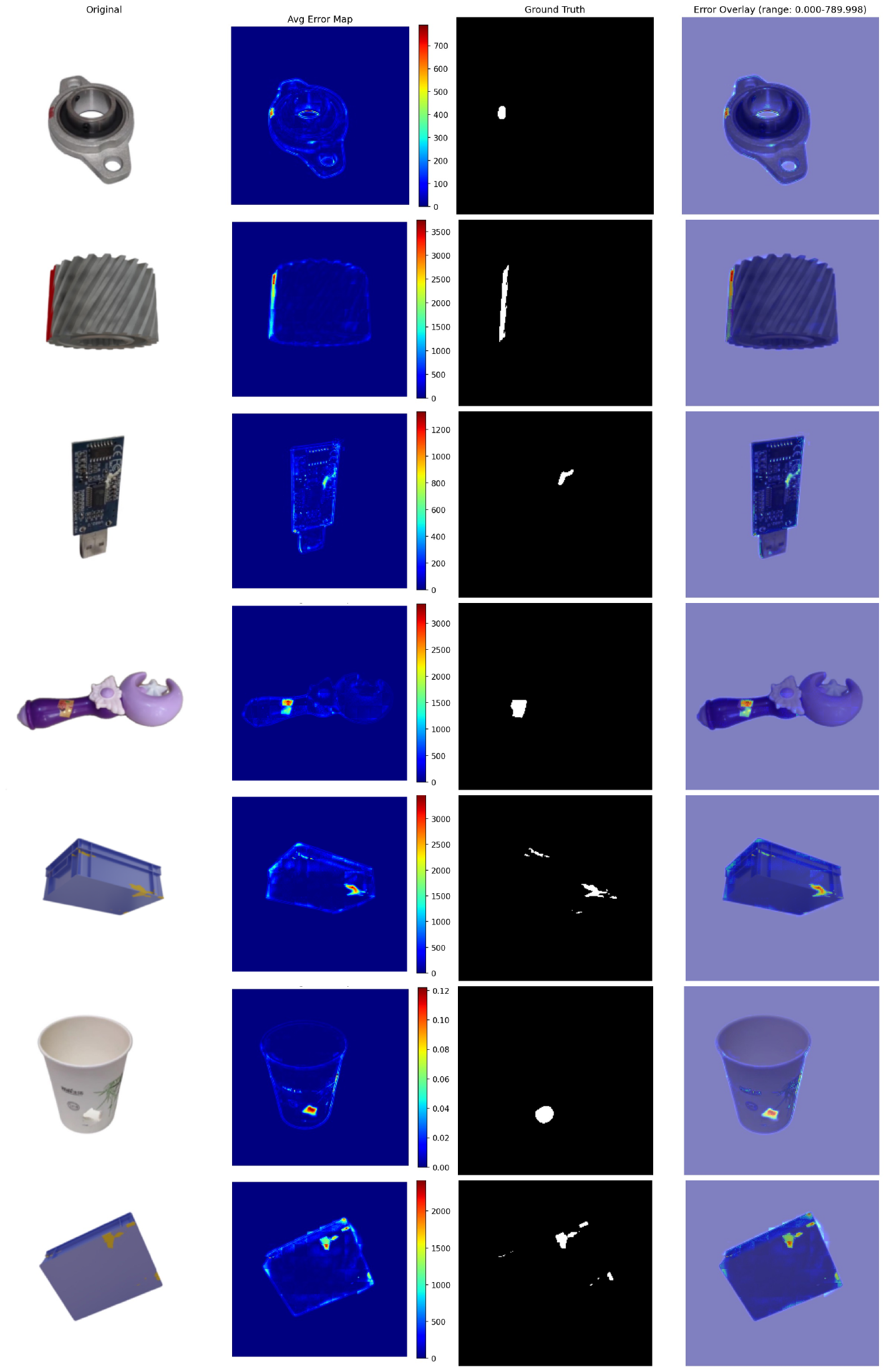}
    \caption{\textbf{Qualitative results - }PIAD 4-shot setup. Our method can generalize on different lighting conditions.}
    \label{fig:supp-qual-piad}
\end{figure*}

\begin{figure*}[t]
    \centering
    \includegraphics[width=\linewidth]{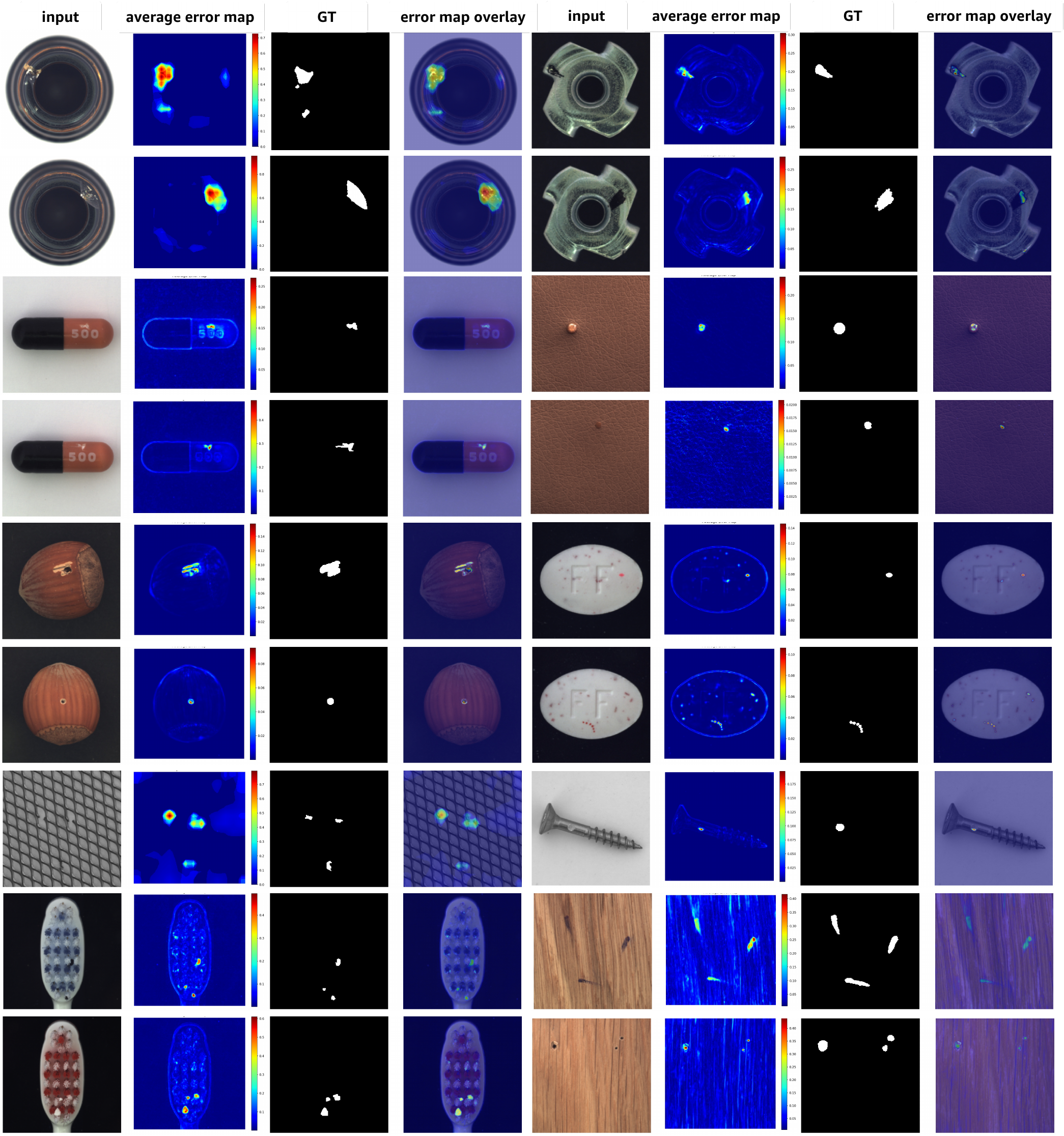}
    \caption{\textbf{Qualitative results - }MVTecAD 4-shot setup. On FSAD task, our method can perform fine-grained anomaly detection well.}
    \label{fig:supp-qual-mvtec}
\end{figure*}

\begin{figure*}[t]
    \centering
    \includegraphics[width=\linewidth]{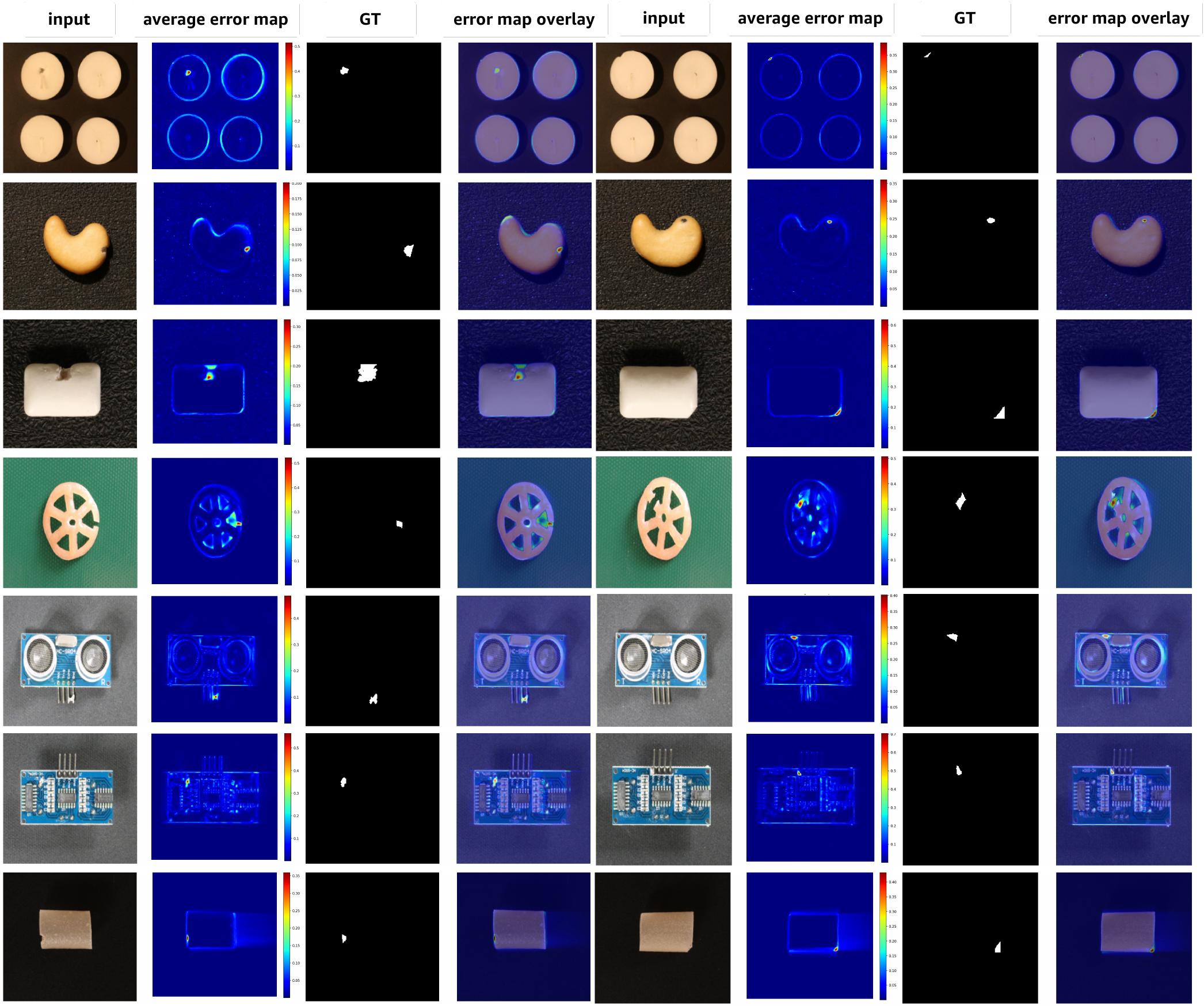}
    \caption{\textbf{Qualitative results - }VisA 4-shot setup. On some challenging cases (row 4), our method may include some noisy predictions due to the reconstruction errors. While it remains effective on very detailed anomalies (row 5, 6).}
    \label{fig:supp-qual-visa}
\end{figure*}

We first show an example of 15 inference runs and their average error map of MAD-Sim with 4-shot set up in Fig.~\ref{fig:supp-qual-1}. Although some inference runs do not masked out the anomaly part or cannot reconstruct an anomaly-free input image successfully, these errors would be eliminated in the final average processing. We further provide qualitative results on MAD-Sim (Fig.~\ref{fig:supp-qual-mad}), PIAD (Fig.~\ref{fig:supp-qual-piad}), MVTecAD (Fig.~\ref{fig:supp-qual-mvtec}) and VisA (Fig.~\ref{fig:supp-qual-visa}).

\end{document}